# An Independent Discriminant Network Towards Identification of Counterfeit Images and Videos

Shayantani Kar [A], B. Shresth Bhimrajka [B], Aditya Kumar [C], Sahil Gupta [D],
Sourav Ghosh [E], Subhamita Mukherjee [F], Shauvik Paul [G]

A. Author is with Department of Information Technology, Techno Main Salt Lake, Kolkata 700091, India, ORCID: 0009-0005-9670-118X (e-mail: imshayantani@gmail.com).

B. Author is with Department of Information Technology, Techno Main Salt Lake, Kolkata 700091, India, ORCID: 0009-0003-6717-0442 (e-mail: bhimrajkashresth@gmail.com).

C. Author is with Department of Information Technology, Techno Main Salt Lake, Kolkata 700091, India, ORCID: 0009-0005-9917-6122 (e-mail: adityakr0004@gmail.com).

D. Author is with Department of Information Technology, Techno Main Salt Lake, Kolkata 700091, India, ORCID: 0009-0000-6931-1008 (e-mail: sahilgupta9700@gmail.com).

E. Author ORCID – 0000-0003-1866-1408 (e-mail: Sourav.ghosh@outlook.in)

F. Author is with Department of Information Technology, Techno Main Salt Lake, Kolkata 700091, India, ORCID: 0000-0003-3924-5833 (Phone: +919830370605; e-mail: subhamita.mukherjee@gmail.com)

G. Author is with Department of Information Technology, Techno Main Salt Lake, Kolkata 700091, India, ORCID: 0009-0008-3377-8448 (e-mail: paulshauvik@gmail.com)

**ABSTRACT**

Rapid spread of false images and videos on online platforms is an emerging problem. Anyone may add, delete, clone or modify people and entities from an image using various editing software which are readily available. This generates false and misleading proof to hide the crime. Now-a-days, these false and counterfeit images and videos are flooding on the internet. These spread false information. Many methods are available in literature for detecting those counterfeit contents but new methods of counterfeiting are also evolving. Generative Adversarial Networks (GAN) are observed to be one effective method as it modifies the context and definition of images producing plausible results via image-to-image translation. This work uses an independent discriminant network that can identify GAN generated image or video. A discriminant network has been created using a convolutional neural network based on InceptionResNetV2. The article also proposes a platform where users can detect forged images and videos. This proposed work has the potential to help the forensics domain to detect counterfeit videos and hidden criminal evidence towards the identification of criminal activities.

## 1. INTRODUCTION

Generative adversarial networks (GAN) is an advanced generative modelling technique under deep learning which uses convolutional neural networks (CNN) [20]. Generative Modelling is an unsupervised learning method where the model can automatically train itself by learning and discovering a certain pattern from the given data. Then the model can output probable results. These outputs are deceiving as they look realistic. Advanced GANs make it more difficult for humans to identify difference between generated image and real image. The GANs are a method of training the generative model with the help of two sub-models– generator and discriminator. They compete each other to give better results. The generator produces new processed frames and the discriminator model classifies the produced frames as genuine or counterfeit. These two models are trained recursively until the discriminator fails to identify a counterfeit example, which means the generator can produce feasible examples.

Figure 1(b) is a GAN generated counterfeit version of figure 1(a). This counterfeit image is nearly impossible to perceive through human eye. GANs are rapidly getting powerful on the basis of generative models. They can deliver generated realistic images for a vast range of problem domains, mostly image-to-image translation [3].

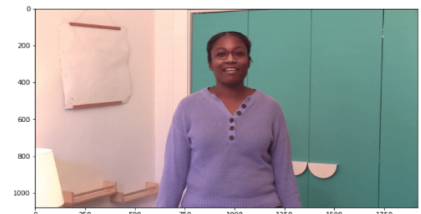

**Figure 1(a).** An original image

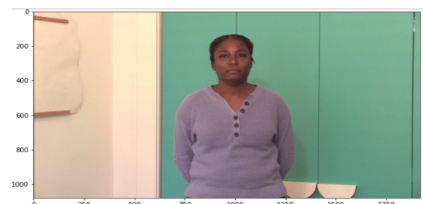

**Figure 1(b).** A GAN generated image

Now-a-days, image creation is predominant use of GAN. Existing photo editing software is capable of image editing from scratch but those have inherent limitation. Those are also not capable of producing deceiving images. GAN can generate deceiving images. These deceiving images have potential threat to spread false information.

Frauds are using GAN in a negative context to cover video or image proofs. It uses synthetic video or image content for committing various illicit activities, including blackmail, cybercrime, social engineering, phishing etc. People are also using social media platforms to spread misleading counterfeit information with edited images to portray a crafted counterfeit incident. Social platforms contain many such counterfeit images and videos which create negative impact.

This article proposes a discriminant network to identify difference between real and processed counterfeit images and videos with the help of convolutional neural networks (CNN). This work has modified InceptionResNetV2 to create a model which can detect counterfeit images and each frame of videos. It extracts frames from a video to train the model, and then feed it with several GAN generated images to gain high accuracy and precision in the result. The work also introduces a mobile application to identify counterfeit images and expose edited deceptive criminal evidences. Finally, it has discussed a possible system architecture for integrating the application on cloud platform.

Rest of the article is organized with literature survey in next section followed by proposed methodology is section 3. Experimental results and performance analysis are discussed in section 4 and conclusion in section 5.

## 2. Literature Survey

Easy access to internet and its rapidly growing use has induced the culture of illicit activities with edited media contents including image and video. GAN is a deep learning model that generates samples from statistical distribution of data which are beyond the human perception level. This has a negative impact during forensic investigation of criminal activities. An analysis has been performed in [23] on forensic detection process to differentiate between GAN manufactured images and photographed images. The authors have discussed potential risks in such detection processes.

A study to identify gap between counterfeit and camera images has measured people's ability to perform visual inspection on edited images [1]. Survey report shows that, only 62% to 66% of photos were correctly classified and users found it very challenging to identify the manipulation. Thus, the threat posed by large-scale image manipulation has been an alarming concern in media forensics.

Another work visualized computer image based on wavelet decomposition [5] or residual image patterns [6]. Other studies investigate on audio differences introduced by the recording device [7], and traces of chromatic aberrations [8]. A work in [10] has focused on the differences between colour distribution. Statistical properties of local fringe fields for discrimination are used in [11]. A facial asymmetry analysis is proposed in [12] as a discriminative feature to distinguish computer-generated from natural human faces.

Use of deep learning models in this domain are found in [4], [13], [14] and the performances of deep learning models are encouraging than previous approaches. DeepFake images are classified using saturation cues in [18]. A structure of generating networks from one of the significant GAN implementations [19] reveals that the exposure of the network can compromise the images. Another work in [20] has explored social networks using 36302 images based on XceptionNet and achieved an accuracy of 89% on compressed data using deep learning model. A temporal-aware pipeline to detect DeepFake videos is discussed in [22]. Deep forgery discriminator (DeepFD) was developed in [21] using contrastive loss function to efficiently detect DeepFake images. Different methods to discriminate GAN generated images using various algorithms like self-attention mechanism, incremental learning, Benford's law, two-step pair wise learning have been discussed in [24], [25], [26] and [27] respectively. Advancement of graphics software has left immense challenge for researchers to identify between processed and photographed images [15].

Recent researches are focusing on many advanced optimization algorithms on diverse field like depth-information based differential evolution mechanism [28], artificial rabbit optimization [29], artificial hummingbird algorithm [30], co-operative strategy based differential evolution [31], non-linear finite element model [32], different machine learning models [33], convolutional neural network with radial reinforcement fuzzy model [34] and quick crisscross sine cosine algorithm [35] and many more.

The proposed model in this article aims to deploy an independent discriminant network that can identify a GAN generated image using InceptionResNet model.

## 3. Proposed Methodology

Many people have inhibited growing tendency to counterfeit different images on the internet through image editing software. They use different filter and noise to create real-alike counterfeit images. Generative adversarial network (GAN) is a popular method of generating images from scratch on basis of set of text or graphical inputs. Pipelines using GAN comprises a discriminant network that learns to perform a binary classification on images to identify if the video frame or image is manufactured or real. The proposed work aims for training the model using images manufactured by various GANs. An algorithm has been designed that can independently and accurately classify the image in real or manufactured category. A cloud-based application has been discussed in the next phase which helps to separate artificially generated images and frames of videos from the real ones. This application has two major phases; first one is a machine learning model and the next phase builds application with cloud integration.

## 3.1 Phase 1 – Machine Learning

A large volume of training data has been gathered in this step to train a model. A machine learning model is chosen after systematic review. It has identified InceptionResNet model. This model is fusion of Inception networks and ResNet networks. Figure 2 represents the flowchart of this process.

### 3.1.1 Step 1. Data visualization and Analysis

Data is cleaned, visualized and analysed in this phase. It prepares and cleans the raw data to make it useable by the model. It performs face detection in video and crop out the required section as training images. MMOD (Max Margin Object Detection) is used for this step. Parameters for facial detection are,

a. 2x scale for pixels < 300
b. No scale for 100 < pixels < 300
c. 0.5x scale for pixels > 1000
d. 0.33x scale for pixels > 1900

These constraints are applied so that the images retain similar size as the native video dataset of 340 x 340 pixels.

### 3.1.2 Step 2. Facial detection using an MMOD based CNN

The first step of Data Augmentation is to extract facial images from each of the videos to utilize them in the model as inputs. Dlib library is utilized for this purpose which uses a Max Margin Object Detection (MMOD) model with Convolutional Neural Networks (CNN). All videos are loaded and start extracting only 10 frames from each video to save space. The resultant output is stored in a dataset folder segregated according to real or counterfeit values.

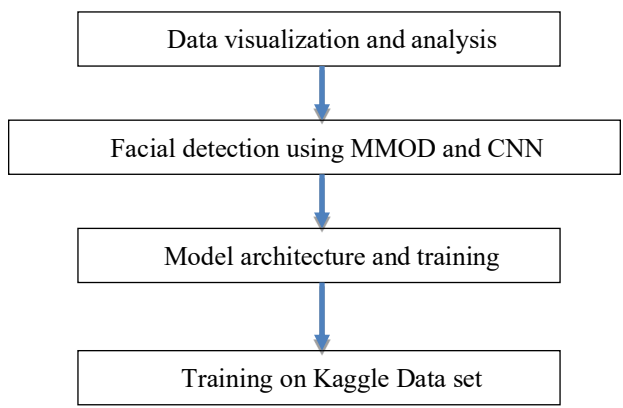

```
Data visualization and analysis
        ↓
Facial detection using MMOD and CNN
        ↓
Model architecture and training
        ↓
Training on Kaggle Data set
```

**Figure 2.** Flowchart of the proposed method in phase 1

### 3.1.3 Step 3. Model Architecture and Training

#### 3.1.3.1 Declaring Constants

Image size is set as 256 x 256 shape with RGB colour values. Also training batch of size 32 is used.

#### 3.1.3.2 Training Pipeline

The entire dataset cannot be fit directly on the memory because of the huge size of dataset so a data pipeline is utilized to feed the data for training the model. The TensorFlow Data API based tools are needed to build the pipeline for training, validation, and data testing. The pipeline reads the data from the disk sequentially and uses mapping functions to convert the data into suitable forms to be processed by the model. The mapped values can then be cached to reduce I/O times. Parallel processing is utilized for better utilization of the CPU to prepare the next training batch while the GPU keeps itself busy training the model.

#### 3.1.3.3 Model Architecture

##### 3.1.3.3.1 EfficientNetB0

EfficientNet-B0 is a convolutional neural network that has been trained to classify images into 1000 object types, such as trees, computer, and even human faces. Thus, rich point representations have been learned by the network for a large volume of images. The network has an input of images of size 224x224. It slightly scales network range, resolution, and depth with fixed scaling portions, but conventional practice randomly fixes these values. If the size of input image increases it requires more channels to catch fine-pixelated patterns on the larger image. The network also requires more layers.

It follows the semi-supervised literacy approach which states that:

a. EfficientNetB0 model is first trained on labelled images, then it is utilized to induce unlabelled images.
b. Next, a larger EfficientNetB0 model is trained on the combination of labelled and mock images.
c. These steps are then done multiple times. Noise is also fitted during the process similar as Powerhouse, Stochastic Depth and Data Augmentation.

##### 3.1.3.3.2 ResNet50

Training a new neural network becomes more and more challenging with increasing number of the layers. Importance of ResNet is high under such scenario as it helps to train deeper neural networks efficiently. ResNet stands for Residual Network. It is a type of a convolutional neural network which uses a depth of 50 layers. It uses an input of images of size 224 x 224.

##### 3.1.3.3.3 InceptionResNetV2

A key component of the ML model contains a dense convolutional neural network called InceptionResNetV2. Composed of a conglomeration of multiple starting blocks and residual connections, the overall network is very deep and highly efficient for learning different features quickly. Thus, a model-wide feature detection block was used. The output features produced by the feature detector are fed into a dense layer, giving a single neuron output as the statistical probability that the image is real. It is a convolutional neural network trained on over 1 million images. The network has a depth of 164 layers and can segregate images into 1000 objects, So, the network has been trained to learn functional representation of large number of images. 299 x 299 is the input image size of the network and the result is a list containing the estimation of class probabilities. It is based on a conglomeration of starting structures and remaining connections. The inception ResNet block combines convolutional filters of multiple sizes with the rest of the connections. Using residual connections reduces the training period and reduces the degradation problem caused by deep structures.

The advantages of the InceptionResNet are:

   a.  High power gain in convolutional neural networks

   b.  Ability to extract features from input data at different scales using different convolution filter sizes.

   c.  1x1 convolutional filters learn cross-channel patterns and contribute to global feature extraction network capabilities.

Figure 3 represents the InceptionResNetV2 diagram.

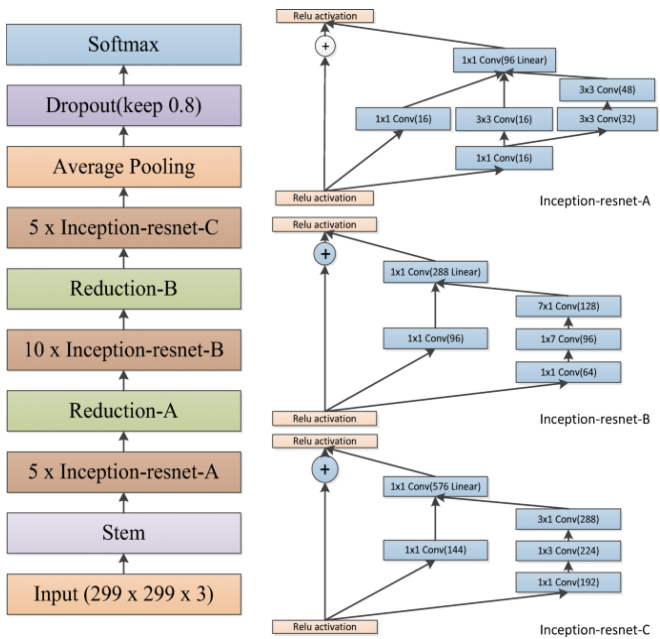

**Figure 3.** Schema for InceptionResNetV2 networks, The schema for 35x35 grid (InceptionResNet-A) module of the InceptionResNetV2 network, The schema for 17X17 grid (InceptionResNet-B) module of the InceptionResNetV2 network, The schema for 8X8 grid (InceptionResNet-C) module of the InceptionResNetV2 network [17]

A dense hidden layer of an artificial neural network (ANN) is merged with InceptionResNetV2. This is used in activations based on sigmoid functions. The sigmoid function has a unique characteristic S-shaped curve as represented in Figure 4(a). A well-known sigmoid function is the logistic function. It is also called as the logistic sigmoid function and is often used in statistical analysis and machine learning. The entire number line is mapped onto a small region (usually between 0 and 1, or -1 and 1). Here the sigmoid function can convert real values to probabilities.

The final level of the model follows the SoftMax activation function. SoftMax functions are powerful tools in mathematics and data analysis. This has become more and more popular in recent years. This function considers a vector of K real numbers and transforms it to a probability distribution of K possible outcomes as shown in Figure 4(b). This is achieved through a process of normalization and augmentation, resulting in a versatile and flexible tool for a wide range of applications.

$$Softmax\ \sigma(\vec{z})_i = \frac{e^{z_i}}{\Sigma_{j=1}^{k} e_j^z}$$

$$Sigmoid\ S(x) = \frac{1}{1 + e^{-x}}$$

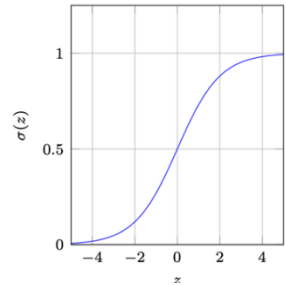
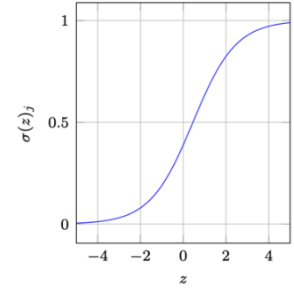

**Figure 4(a).** Sigmoid graph      **Figure 4(b).** **Softmax** graph

### 3.1.4 Step 4. Training on Kaggle Dataset

Once the model is prepared, process is initiated for training the dataset extracted from Kaggle. The model's performance is evaluated thorough sampling of a set containing 128 test images and using the model for prediction of the labels. The predictions are then mapped into a confusion matrix to evaluate other metrics.

## 3.2 Phase 2 – Application Development

A mobile application-based frontend is prepared after deploying the model. This application provides an user interface for end users.

### 3.2.1 Frontend Development

The frontend is developed using React Native on an Expo environment. React Native uses a framework as it helps to build a platform independent application which runs on android, iOS, and web. Basically, this is a JavaScript programming language. Expo provides a development environment for building, testing, deployment, and maintenance of applications.

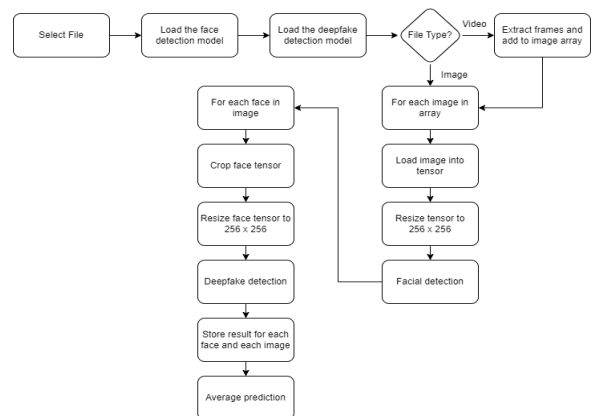

**Figure 5.** Workflow of the Backend System

The main modules and functionalities of the application are;

a. Uploading both Image and Videos to check for computer manipulation.
b. Detailed information on each face detected whether real or not.
c. No images or video data is collected so privacy is maintained.

### 3.2.2 Backend Development

The backend of the application handles all the major steps for predicting whether the image or video is real or processed as shown in Figure 5. The backend system provides an API that allows the frontend to communicate the requests with the image and videos to be identified. This API is developed by the Flask framework in Python. Python is the most appropriate choice as it provides frameworks for both the API and ML. The trained InceptionResNet model is exported as .pkl file to be utilized in the backend system. The flow of process in the backend is as follows:

a. The API receives a request from the frontend with the payload data which is either an image or a video.
b. The data's file type is determined. If the type is video then the individual frames from the video is extracted by random sampling and returns an image array, else it returns a single element array of the image.
c. For each image in the array, it is passed through the facial detection model for generating an array of face locations.
d. For each face in the face array, it is cropped and passed through the InceptionResNet model to identify if it is counterfeit or not.
e. The average prediction for all faces is stored as overall result and returns a response to the frontend.

This backend system is suitable deployed on a Platform as a Service (PaaS) to be hosted. A PaaS is chosen as it handles all the infrastructural requirements without any intervention. It also provides a development environment suitable for Python applications.

### 3.3 Phase 3 – Cloud Architecture

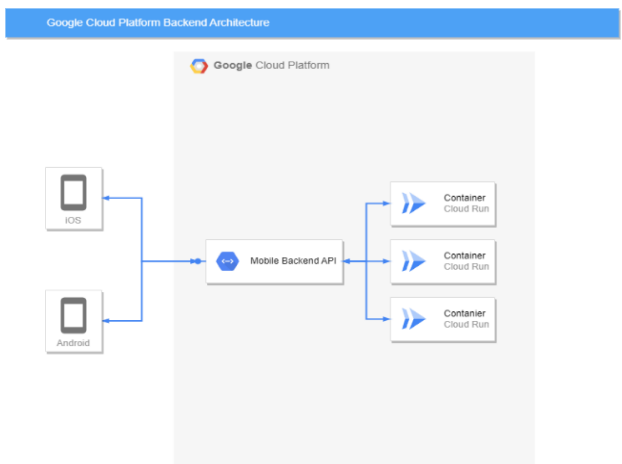

**Figure 6.** Workflow of the Backend System

The backend for this application is deployed on Google Cloud Platform using their Platform as a Service (PaaS) Cloud Run. This service is chosen because of low-cost option which runs the backend only when a request is assessed.

The backend is exported as a docker image which runs as containers on Cloud Run as shown in Figure 6. Cloud Run provides high scalability up to 100 concurrent containers and traffic management which avoids the need of using an application such as Kubernetes

## 4. EXPERIMENTAL DETAILS

The Discriminant Network is developed by using three Deep Learning – Convolutional Neural Networks

a. ResNet50
b. EfficientNetB0
c. InceptionResNetV2

At first, the data is processed and analysed, Figure 7 shows the results:

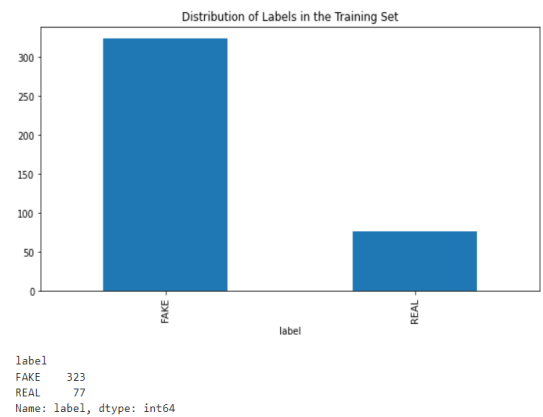

**Figure 7.** Analysis of the Metadata

**Figure 8.** Analysis of the distribution of the data

Figure 7 and 8 describe analysis of the metadata of the dataset and analysis of the distribution of the data.

Two different types of datasets have been used so as to train and test the Discriminant network on two separate GAN generated datasets. The datasets are –

a. Nvidia Dataset – A dataset generated with a powerful GAN model.
b. Kaggle Dataset – A basic dataset generated by weaker GAN models.

Post processing the data is trained with three different Deep Learning algorithms to get the results. The results and Accuracy obtained using these 3 networks are shown below:

## 4.1 ResNet50

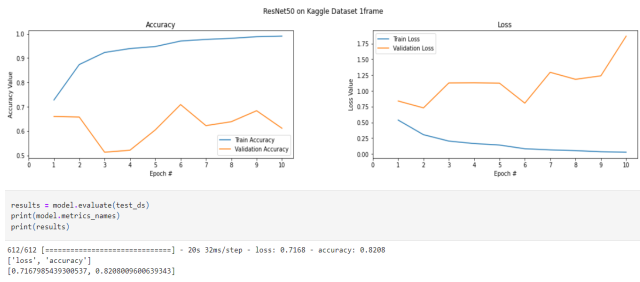

**Figure 9.** Accuracy obtained using ResNet50 in Kaggle dataset

It is observed in Figure 9 that the model was not suitable enough to handle other more powerful GAN generated images that were present in the Nvidia dataset.

## 4.2 EfficientNetB0

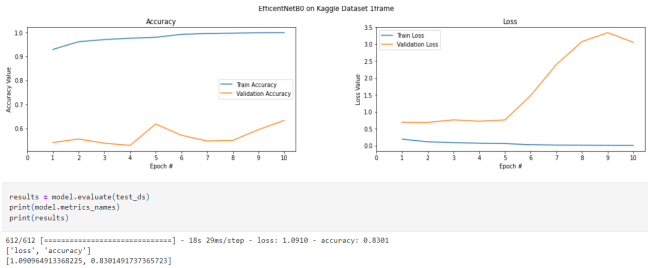

**Figure 10.** Accuracy obtained using EfficientNetB0 in Kaggle dataset

It is observed in Figure 10 that the model was not suitable enough to handle other more powerful GAN generated images that were present in the Nvidia dataset.

## 4.3 InceptionResNetV2

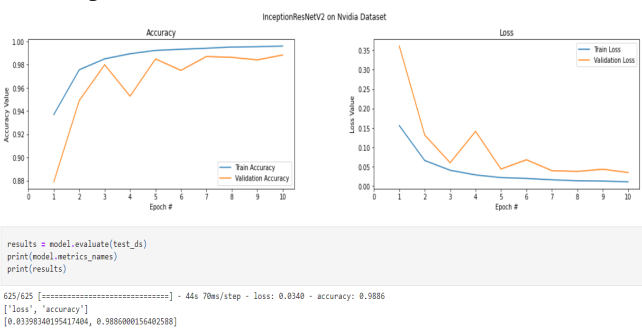

**Figure 11.** Accuracy obtained using InceptionResNetV2 in Kaggle dataset

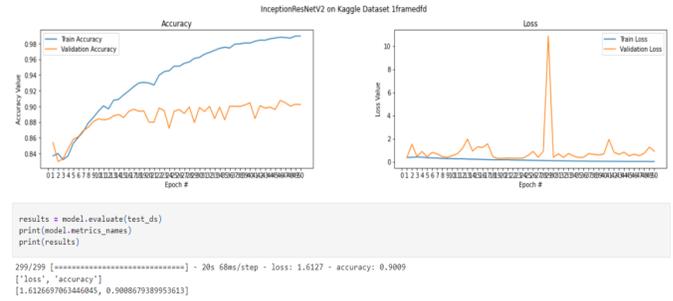

**Figure 12.** Accuracy obtained using InceptionResNetV2 in Kaggle dataset

It is observed in Figure 11 and Figure 12 that the InceptionResNetV2 model gives the best accuracy on both Nvidia and Kaggle dataset with an accuracy of 90.08%.

The detailed performance analysis is discussed in the following section.

## 4.4 Performance Analysis

After training the model using three CNN algorithms, we study the comparative analysis of the results given by them based on the Kaggle Dataset.

Table 1: Comparative Analysis of the Models

| Parameters | ResNet50 | EfficientNetB0 | InceptionResNetV2 |
|---|---|---|---|
| True Positive | 106 | 108 | 120 |
| False Positive | 22 | 20 | 87 |
| True Negative | 103 | 104 | 83 |
| False Negative | 25 | 24 | 45 |
| F-score | 0.8185328 | 0.8307692 | 0.819112 |
| Precision | 0.828125 | 0.843750 | 0.93750 |
| Recall | 0.80916 | 0.818182 | 0.727273 |
| Accuracy | 82.081% | 83.01% | **90.08%** |

The model's performance is evaluated by randomly sampling a set of 128 test images and using the model to predict the labels. The predictions are then mapped into a confusion matrix to evaluate other metrics.

The comparative analysis in Table 1 shows that InceptionResNetV2 gives much better accuracy in both the datasets, which is graphically represented in Figure 13.

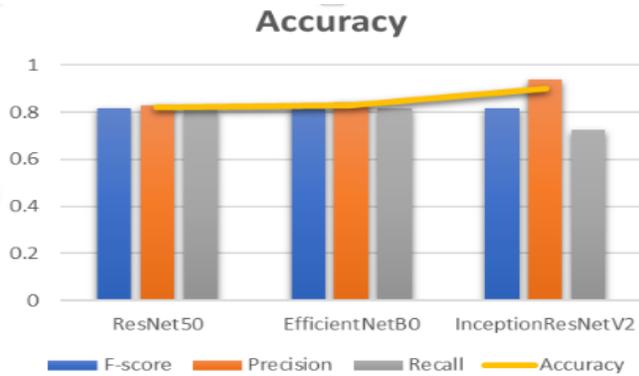

**Figure 13.** Comparative analysis of the models

### 4.5 User Interface Design of the Application

Figure 14(a) and 14(b) shows the user interface that has been developed using React Native and Expo. The proposed frontend architecture has been designed taking into account user-friendliness and clarity of output. Here are few examples of the interface:

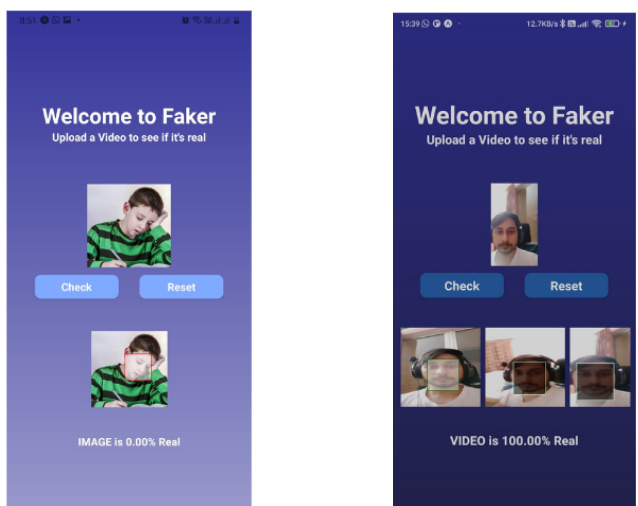

**Figure 14.** (a) The user interface showing a GAN generated image as counterfeit (b) The user interface showing a video as real extracting each frame of the video.

## 5. Conclusion

Rapid spread of DeepFake images is an emerging concern. A modern technology like GANs are getting heavily used across internet for different counterfeiting of images and videos. This counterfeiting is affecting art, entertainment, education and many more sectors. DeepFakes are also frequently used for harassment, intimidation and extortion against individuals and businesses. There are many models to identify these DeeFakings but intruders are adopting new policy to continue the abuse. Deep learning-based models available in the literature show potential improvement in the identification of DeepFaking. This proposed model using independent discriminant network towards identification of counterfeit images and videos can potentially put up better results. The mobile application which is integrated with the proposed independent discriminant network-based model, has made the task of identification easier for the users. The task has every potential to help the forensics and can safeguard against cybercrime.

**CONFLICT OF INTEREST**

The authors declare that they have no conflict of interest.

**DATA AVAILABILITY STATEMENT**

Data sharing not applicable to this article as no datasets were generated or analysed during the current study.

## Acknowledgment

The authors of this paper thankfully acknowledge the teaching fraternity of the Department of Information Technology for the help and support during this research work.

## References

1. A. Dirik, S. Bayram, H. Sencar, and N. Memon, "New features to identify computer generated images," in IEEEICIP, Oct 2006.
2. J.-F. Lalonde and A. Efros, "Using color compatibility for assessing image realism," in IEEE ICCV, Oct 2007.
3. A. Gallagher and T. Chen, "Image authentication by detecting traces of demos icing," in IEEE CVPR Workshops, June 2008.
4. S. Lyu and H. Farid, "How realistic is photorealistic?" IEEE transactions on Signal Processing, vol. 53, no. 2. 2009.
5. Davis E. King. Dlib-ml: A Machine Learning Toolkit. Journal of Machine Learning Research, 2009.
6. S. Dehnie, H. Sencar, and N. Memon, "Digital image forensics for identifying computer generated and digital camera images," in IEEE ICIP, 2010.
7. R. Wu, X. Li, and B. Yang, "Identifying computer generated graphics via histogram features," in IEEE ICIP, 2011.
8. R. Zhang, R.-D. Wang, and T.-T. Ng, "Distinguishing photographic images and photorealistic computer graphics using visual vocabulary on local image edges," in International Workshop on Digital Forensics and Watermarking, Oct 2011.
9. D.-T. Dang-Nguyen, G. Boato, and F. D. Natale, "Discrimination between computer generated and natural human faces based on asymmetry information," in Eusipco, 2012.
10. D. Cozzolino, D. Gragnaniello, and L.Verdoliva, "Image forgery detection through residual-based local descriptors and block-matching," in IEEE Conference on Image Processing(ICIP), October 2014
11. O. Holmes, M. Banks, and H. Farid, "Assessing and improving the identification of computer-generated portraits," 2016.
12. S. Nightingale, K. Wade, and D. Watson, "Can people identify original and manipulated photos of real-world scenes? Cognitive Research: Principles and Implications," 2017.
13. V. Schetinger, M. Oliveira, R. da Silva, and T. Carvalho,"Humans are easily fooled by digital images," Computers &Graphics, vol. 68, 2017.

14. S. Fan, T.-T. Ng, B. Koenig, J. Herberg, M. Jiang, Z. Shen,and Q. Zhao, "Image visual realism: From human perception to machine computation," IEEE Transactions on Pattern Analysis and Machine Intelligence, in press, 2017.

15. N. Rahmouni, V. Nozick, J. Yamagishi, and I. Echizeny,"Distinguishing computer graphics from natural images using convolutional neural networks," in IEEE WIFS, 2017.

16. E. de Rezende, G. Ruppert, and T. Carvalho, "Detecting computer generated images with deep convolutional neural networks," in SIBGRAPI Conference on Graphics, Patterns and Images, 201.

17. Peng, C., Liu, Y., Yuan, X. et al. Research of image recognition method based on enhanced inception-ResNet-V2. Multimed Tools Appl 81, 34345–34365 (2022).

18. S. McCloskey and M. Albright, "Detecting GAN-Generated Imagery Using Saturation Cues," 2019 IEEE International Conference on Image Processing (ICIP), Taipei, Taiwan, 2019, pp. 4584-4588, doi: 10.1109/ICIP.2019.8803661.

19. Tero Karras, Timo Aila, Samuli Laine and Jaakko Lehtinen, "Progressive growing of GANs for improved quality stability and variation", International Conference on Learning Representations, 2018.

20. F. Marra, D. Gragnaniello, D. Cozzolino and L. Verdoliva, "Detection of gan-generated counterfeit images over social networks", 2018 IEEE Conference on Multimedia Information Processing and Retrieval (MIPR), 2018.

21. C.-C. Hsu, C.-Y. Lee and Y.-X. Zhuang, "Learning to Detect Counterfeit Face Images in the Wild", ArXiv e-prints, Sept. 2018.

22. D. Guera and E. J. Delp, "DeepFake video detection using recurrent neural networks", IEEE International Conference on Advanced Video and Signal-based Surveillance (to appear), 2018.

23. H. Li, H. Chen, B. Li and S. Tan, "Can Forensic Detectors Identify GAN Generated Images?," 2018 Asia-Pacific Signal and Information Processing Association Annual Summit and Conference (APSIPA ASC), Honolulu, HI, USA, 2018, pp. 722-727, doi: 10.23919/APSIPA.2018.8659461.

24. Z. Mi, X. Jiang, T. Sun and K. Xu, "GAN-Generated Image Detection With Self-Attention Mechanism Against GAN Generator Defect," in IEEE Journal of Selected Topics in Signal Processing, vol. 14, no. 5, pp. 969-981, Aug. 2020, doi: 10.1109/JSTSP.2020.2994523.

25. F. Marra, C. Saltori, G. Boato and L. Verdoliva, "Incremental learning for the detection and classification of GAN-generated images," 2019 IEEE International Workshop on Information Forensics and Security (WIFS), Delft, Netherlands, 2019, pp. 1-6, doi: 10.1109/WIFS47025.2019.9035099.

26. N. Bonettini, P. Bestagini, S. Milani and S. Tubaro, "On the use of Benford's law to discriminate GAN-generated images," 2020 25th International Conference on Pattern Recognition (ICPR), Milan, Italy, 2021, pp. 5495-5502, doi: 10.1109/ICPR48806.2021.9412944.

27. Y. -X. Zhuang and C. -C. Hsu, "Detecting Generated Image Based on a Coupled Network with Two-Step Pairwise Learning," 2019 IEEE International Conference on Image Processing (ICIP), Taipei, Taiwan, 2019, pp. 3212-3216, doi: 10.1109/ICIP.2019.8803464.

28. P. Jangir, A.E. Ezugwu, Arpita et al. "Precision parameter estimation in Proton Exchange Membrane Fuel Cells using depth information enhanced Differential Evolution", Sci Rep 14, 29591 (2024).

29. P. Jangir, A.E. Ezugwu, K. Saleem et al. "A hybrid mutational Northern Goshawk and elite opposition learning artificial rabbits optimizer for PEMFC parameter estimation". Sci Rep 14, 28657 (2024).

30. P. Jangir, A.E. Ezugwu, K. Saleem et al. "A levy chaotic horizontal vertical crossover based artificial hummingbird algorithm for precise PEMFC parameter estimation". Sci Rep 14, 29597 (2024).

31. P. Jangir, Arpita, S.K. Agrawal et al., "A cooperative strategy-based differential evolution algorithm for robust PEM fuel cell parameter estimation". Ionics. 31, 2024, 703-741. 10.1007/s11581-024-05963-x.

32. H.S. Mohamed, T., Qiong, H.F. Isleem et al. "Compressive behavior of elliptical concrete-filled steel tubular short columns using numerical investigation and machine learning techniques" Sci Rep 14, 27007 (2024).

33. Y. Chen, F. Haytham, D. N. Isleem et al., "Utilization finite element and machine learning methods to investigation the axial compressive behavior of elliptical FRP-confined concrete columns", Structures, Volume 70, 2024, 107681, ISSN 2352-0124,

34. G. Srikanth, D. Nimma, RVS Lalitha et al. "Food Security Based Marine Life Ecosystem for Polar Region Conditioning: Remote Sensing Analysis with Machine Learning Model", Remote Sens Earth Syst Sci (2024).

35. Sunilkumar, P. Agrawal, P. Jangir, et al., "The quick crisscross sine cosine algorithm for optimal FACTS placement in uncertain wind integrated scenario based power systems", Results in Engineering, Volume 25, 2025, 103703, ISSN 2590-1230

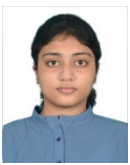

**Shayantani Kar** has pursued her B.Tech. degree in Information Technology from Techno Main Salt Lake (Maulana Abdul Kalam Azad University of Technology), Kolkata, India from 2019-23. She also has a published paper at an international conference. Her areas of interest are Software development, Cloud Computing and Machine Learning.

E-mail: imshayantani@gmail.com

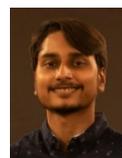

**Shresth Bhimrajka** has pursued his B.Tech. degree in Information Technology from Techno Main Salt Lake (Maulana Abdul Kalam Azad University of Technology), Kolkata, India from 2019-23. He also has a published paper at an international conference. His areas of interests are Software development, Machine Learning, Artificial Intelligence.

E-mail: bhimrajka.shresth@gmail.com

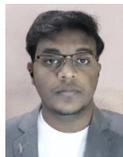

**Aditya Kumar** has pursued his B.Tech. degree in Information Technology from Techno Main Salt Lake (Maulana Abdul Kalam Azad University of Technology), Kolkata, India from 2019-23. He has interned at Informatica as a Software Development Engineer Intern. He also has a published paper at an international conference. His areas of interests are Software development and Machine Learning.

E-mail: adityakr0004@gmail.com

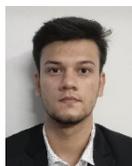

**Sahil Gupta** has pursued his B.Tech. degree in Information Technology from Techno Main Salt Lake (Maulana Abdul Kalam Azad University of Technology), Kolkata, India from 2019-23. He has interned at Informatica as a Software Development Engineer Intern. His areas of interests are Software development and Machine Learning.

E-mail: sahilgupta9700@gmail.com

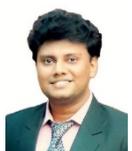

**Sourav Ghosh** received his B.Tech. degree in Information Technology from Techno India (West Bengal University of Technology), Kolkata, India, in 2014, and M.Tech. degree in Computer Science from Indian Institute of Technology, Kharagpur, India, in 2017. He is currently working as a Senior Chief Engineer (AI/ML research) with Samsung R&D Institute Bangalore, India, and apart from his regular industry-academia collaborations, he offers mentorship to students at his alma mater. His current research interests include On-Device AI, Generative AI, Machine Learning, Natural Language Processing, and Computer Vision. He has published more than 13 papers in peer-reviewed international journals and conferences, and has served as a peer reviewer at over 10 top-tier conferences, including ACL, EMNLP, ARR, ACL-IJCNLP, NAACL-HLT, CSAE, ICON, among others.

E-mail: sourav.ghosh@outlook.in

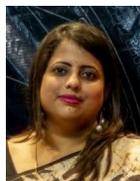

**Subhamita Mukherjee** received her B. Tech. degree in Information Technology from West Bengal University of Technology, India, and M. Tech. degree in Information Technology from IIEST, Shibpur, India. She has completed her Ph.D. in Engineering from IIEST, Shibpur, India. She is currently working as the Associate Professor and Head of the Department of Information Technology at Techno Main Salt Lake, Kolkata, India. Her area of interests are Artificial Intelligence, Algorithm Design and Biochip. She has over 20 published papers in international journals and conferences. She also has 2 published patents.

E-mail: subhamita.mukherjee@gmail.com

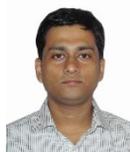

**Shauvik Paul** received his B.Tech. degree in Information Technology from Government College of Engineering and Ceramic Technology, Kolkata, India and M.Tech. degree from Jadavpur University, Kolkata, India. He is currently pursuing his Ph.D. degree from Jadavpur University, Kolkata, India. He is an Assistant Professor at Techno Main Salt Lake, Kolkata, India. He has several published papers in journals and conferences. His area of interests are Bioinformatics, Image Processing, Machine Learning and Internet of Things.

E-mail: paulshauvik@gmail.com